\documentclass[journal]{IEEEtran}
\usepackage{ifpdf}

\usepackage{cite}

\ifCLASSINFOpdf
   \usepackage[pdftex]{graphicx}
	 \graphicspath{{./figures/}}
\else
\fi

\usepackage[cmex10]{amsmath}

\usepackage{algorithmic}

\usepackage{array}

\ifCLASSOPTIONcompsoc
  \usepackage[caption=false,font=normalsize,labelfont=sf,textfont=sf]{subfig}
\else
  \usepackage[caption=false,font=footnotesize]{subfig}
\fi

\usepackage{subfloat}

\usepackage{fixltx2e}

\usepackage{url}

\usepackage{amssymb}
\usepackage{amsthm}
\usepackage{lineno}

\usepackage[usenames]{color}
\usepackage{makeidx}  % allows for indexgeneration 
\usepackage{rotating}
\usepackage{bm}
\usepackage{color}
\usepackage{mathrsfs}
\usepackage{multirow}

\definecolor{MyPurple}{rgb}{.5,0,.5}
\definecolor{RevBlue}{rgb}{0.1,0.1,.6}
\definecolor{RevRed}{rgb}{1.0,0.1,0.1}
\definecolor{RevGreen}{rgb}{0.0,0.6,0.0}
\graphicspath{{./Figures/}}

\newcommand{\figref}[1]{Fig.\ \ref{#1}}
\newcommand{\secref}[1]{Sec.\ \ref{#1}}
\newcommand{\appref}[1]{appendix}
\newcommand{\tabref}[1]{Table \ref{#1}}

\newcommand{\comm}[1]{}

\renewcommand{\b}[1]{\textbf{#1}}
\newcommand{\re}[1]{#1}%{{\color{MyPurple}#1}}
\newcommand{\rev}[1]{#1}%{\color{RevRed}#1}}

\newcommand{\Ncut}{\text{Ncut}}
\newcommand{\cut}{\text{cut}}
\newcommand{\ass}{\text{assoc}}

\newcommand{\fast}{\emph{fast}}

\newcommand{\blockV}[2]{
		\left[\begin{array}{c}
				#1 \\ #2
		\end{array} \right]}

\newcommand{\blockM}[4]{
		\left[\begin{array}{cc}
				#1 & #2 \\ #3 & #4
		\end{array} \right]}

\newcommand{\suchthat}{% 
\mathrel{\ooalign{$\ni$\cr\kern-1pt$-$\kern-6.5pt$-$}}} 

\newcommand\blfootnote[1]{%
  \begingroup
  \renewcommand\thefootnote{}\footnote{#1}%
  \addtocounter{footnote}{-1}%
  \endgroup
}

\DeclareMathOperator*{\argmin}{arg\,min}

\begin{document}
\title{Adaptable Precomputation for Random Walker Image Segmentation and Registration}

\author{ Shawn Andrews and Ghassan Hamarneh \\ 
 %\\
%Medical Image Analysis Lab, School of Computing Science, Simon Fraser University 
%8888 University Drive, Burnaby, B.C., Canada, V5A 1S6
}

% and the end of the author line. i.e. if you had this:

\markboth{}%
{Shell \MakeLowercase{\textit{et al.}}: Bare Demo of IEEEtran.cls for Journals}

%Journal of \LaTeX\ Class Files,~Vol.~11, No.~4, December~2012}%
%{Shell \MakeLowercase{\textit{et al.}}: Bare Demo of IEEEtran.cls for Journals}




\maketitle


\begin{abstract}
The random walker (RW) algorithm is used for both image segmentation and registration, and possesses several useful properties that make it popular in medical imaging, such as being globally optimizable, allowing user interaction, and providing uncertainty information. The RW algorithm defines a weighted graph over an image and uses the graph's Laplacian matrix to regularize its solutions.  This regularization reduces to solving a large system of equations, which may be excessively time consuming in some applications, such as when interacting with a human user. Techniques have been developed that precompute eigenvectors of a Laplacian offline, after image acquisition but before any analysis, in order speed up the RW algorithm online, when segmentation or registration is being performed.  However, precomputation requires certain algorithm parameters be fixed offline, limiting their flexibility. In this paper, we develop techniques to update the precomputed data online when RW parameters are altered. Specifically, we dynamically determine the number of eigenvectors needed for a desired accuracy based on user input, and derive update equations for the eigenvectors when the edge weights or topology of the image graph are changed. We present results demonstrating that our techniques make RW with precomputation much more robust to offline settings while only sacrificing minimal accuracy.
\end{abstract}


% \fi
\IEEEpeerreviewmaketitle

\blfootnote{S. Andrews and G. Hamarneh are with the Medical Image Analysis Lab,
School of Computing Science, Simon Fraser University, 8888 University Drive, Burnaby, B.C., Canada, V5A 1S6. (email: \{sda56, hamarneh\}@sfu.ca)}

\section{Introduction} \label{sec_intro}

Segmentation and registration are crucial tasks in medical image interpretation.  While manual segmentation by an expert is accurate, it is also very time consuming and expensive, making computational techniques designed to aid in segmentation an important field of research \cite{olabarriaga2001interaction,freedman2005interactive}. Such techniques are even more necessary in registration, where manual results are often infeasible to obtain \cite{tang2013medical}. 

\subsection{Discrete Segmentation}

Discrete formulations have become popular for automated image segmentation \cite{boykov2001interactive,couprie2011power,ishikawa2009higher,shi2000normalized,delong2009globally}, with the problem of finding an optimal segmentation cast as minimizing a Markov random field (MRF) energy, allowing powerful techniques from discrete optimization to be leveraged. The random walker (RW) algorithm is a popular, efficient, and flexible discrete technique that was initially developed by Grady for image segmentation (RWIS) \cite{grady2006random}.
 Minimizing the RW MRF reduces to solving a system of equations, with the system matrix given by the \emph{Laplacian} of a weighted graph defined over the image. The Laplacian is a sparse, positive semi-definite matrix (defined in detail in \secref{sec_basic_rw}), thus RWIS is relatively straightforward to implement and provides a globally optimal solution. RWIS provides several other features that make it useful for medical segmentation applications: it extends trivially to multi-label segmentation and higher dimensional images; it provides a probabilistic segmentation, useful for evaluating segmentation uncertainty, which can in turn be used to identify segmentation errors \cite{udupa2002go,saad2010exploration,top2011active}; and it allows user interaction through user specified ``seed'' voxels that are fixed by the user.

Interactive segmentation has been an active research field in medical imaging, due to the difficulty of the segmentation problems and the frequent necessity of expert verified accuracy \cite{olabarriaga2001interaction,freedman2005interactive,boykov2001interactive,falcao1998user,top2011active}. While the listed properties make RWIS a competitive interactive segmentation algorithm, its main drawback is the cost involved in solving the large system of equations between each batch of user input, which can result in time spent waiting for the user, particularly for volumetric images. This issue was greatly mitigated by Grady and Sinop \cite{grady2008fast} by taking advantage of the time an image is available \emph{offline} (after acquisition but before any user interaction) to precompute eigenvector/value pairs of the Laplacian that can greatly speed up the RW algorithm \emph{online} (when a user is actively interacting with the image). The number of eigenvectors controls the trade-off between speed and accuracy, and it was shown that a significant speed-up can be achieved with a minimal loss in accuracy. This technique was extended by Andrews et al. \cite{andrews2010fast} to apply even when  prior information was not available offline \cite{grady2005multilabel}. We refer to this technique as \fast RWIS.

\subsection{Discrete Registration}

The success of discrete techniques for image segmentation has lead to similar discrete formulations for registration \cite{glocker2008dense, tang2007non,heinrich2012globally}. Some discrete registration techniques are based on multi-label optimization techniques, with labels corresponding to displacement vectors from a discrete pre-defined set. An example of such a technique is the RW image registration (RWIR) technique developed by Cobzas and Sen \cite{cobzas2011random}. As a globally optimal deformable registration framework, RWIR has been growing in popularity, with multiple recent extensions \cite{popuri2013variational,tang2013random,tang2014reducing, andrews2014isometric}. As with RWIS, the uncertainty information provided by RWIR has been utilized to identify registration error \cite{lotfi2013improving}. 

While RWIR permits user interaction through the placement of landmark points, it is more often guided by prior probabilities for the displacements derived from image similarity terms. Andrews et al. demonstrated that the precomputation techniques applied to RWIS could be utilized to increase the RWIR speed even when no user interaction is used \cite{andrews2014fast}. The key idea is to re-define ``offline time'' as the time when only one of the images being registered is available, which can often be a significantly long period (e.g. when one image is an atlas). Precomputation performed offline on the available image can then be used to speed up RWIR online, when other images become available and the prior probabilities for each displacement vector can be calculated. We refer to this technique as \fast RWIR.

\subsection{Contributions}

While the \fast RW techniques have been shown to achieve excellent increase in online speed with minimal loss in accuracy, they suffer from one fundamental drawback - they rely on data computed offline, when information regarding the problem to be solved is incomplete, thus limiting the flexibility of the RW methods. Specifically, the \fast RW algorithms require the Laplacian matrix and number of precomputed eigenvector/value pairs to be fixed offline. In this paper, in order to mitigate this drawback, we present techniques for:
\begin{itemize}
	\item Efficient online estimation of the number of eigenvectors required for a desired accuracy. This technique is based on the available online information, specifically, user seeds for RWIS and prior displacement probabilities for RWIR.
	\item Updating the eigenvectors/values when the image graph edge weights are altered online, e.g. to provide stronger regularization across certain image edges. 
	\item Updating the eigenvectors/values when the topological structure of the image graph is altered online. Recent work in discrete registration have shown that image dependent techniques for aggregating graph vertices into super-vertices can greatly improve efficiency \cite{popuri2013variational, tang2014reducing}.
\end{itemize}

The remainder of the paper is laid out as follows. In \secref{sec_method}, we review the RW algorithm and \fast RW extensions, and discuss connections to the normalized cuts segmentation framework \cite{shi2000normalized} that will be relevant when introducing our techniques in \secref{sec_online}. We then demonstrate the practical usefulness of our techniques in \secref{sec_results} and conclude the paper in \secref{sec_conclusion}.

\section{Fast Random Walker Overview} \label{sec_method}

\subsection{Random Walker Algorithm} \label{sec_basic_rw}

Let $J$ be an image over the voxel set $\Omega$, with $|\Omega| = N$.  The RW algorithm assigns each voxel a $K$ length probability vector (i.e. a positive vector that sums to $1$), which we will denote by the $N \times K$ row stochastic matrix $U = [\b{u}^1, \dots, \b{u}^K]$. In RWIS, these probabilities correspond to $K$ different structures in an image. In RWIR, the probabilities correspond to $K$ different potential displacement vectors. 

The RW algorithm consists of defining constraints on $U$ and then regularizing it subject to these constraints. Typical constraints are either soft, consisting of an $n \times K$ matrix $P = [\b{p}^1, \dots, \b{p}^K]$ of prior probabilities that $U$ should be similar to, or hard, consisting of user specified ``seeds'' voxels that have one label fixed to probability $1$. In RWIR, these seeds take the form of landmarks. In order to simplify the notation related to the seed constraints, we re-order the voxels so the rows corresponding to seeded voxels come first, and then divide $U$ into ``\b{s}''eeded and ``\b{n}''on-seeded components using block matrix notation:
\begin{align}
	U &= \blockV{U_s}{U_n}\;, \quad \b{u}^k = \blockV{\b{u}^k_s}{\b{u}^k_n}  \label{eq_U_block} \;,
\end{align}
where $U_s$ is a fixed constant and $U_n$ are the remaining unknowns. We denote $S \ll N$ as the number of seeds.

\begin{figure*}[t]
	\centering
	
	\subfloat[Image]{
		\includegraphics[width=0.14\textwidth]{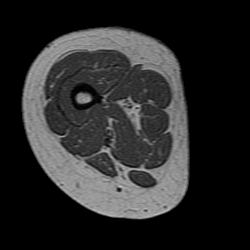}
	}
	\quad
	\subfloat[Eigenvectors]{
		\includegraphics[width=0.14\textwidth]{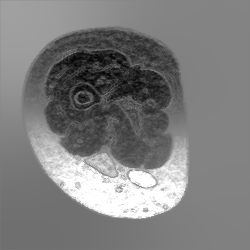} \;
		\includegraphics[width=0.14\textwidth]{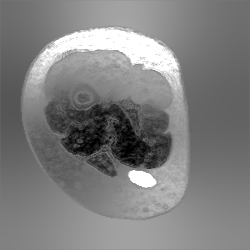} \;
		\includegraphics[width=0.14\textwidth]{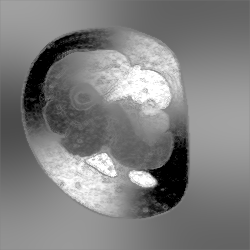} \;
		\includegraphics[width=0.14\textwidth]{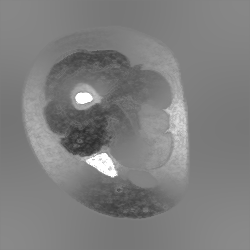} \;
		\includegraphics[width=0.14\textwidth]{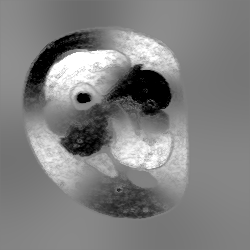}
	} 
	
	\caption{
		A slice of a thigh MRI and several corresponding eigenvectors. The eigenvectors correspond to relaxed normalized cuts of the image, and thresholding them would result in a segmentation of certain structures, albeit perhaps not a useful one, as no user guidance has yet been given.
	}
	\label{fig_eig_ex}
\end{figure*}

In RWIS, the prior probabilities can be calculated from statistics on the user input seeds, allowing spatially disjoint objects to be segmented without seeding each object individually \cite{grady2005multilabel}. In RWIR, the soft constraints are usually based on a local image similarity measure, and are the driving force behind the registration.

The RW regularization is done using an image graph, with vertices corresponding to voxels and weighted edges between neighboring voxels, where the weights control how similar the probabilistic labels of neighboring voxels should be. We define $W$ as the matrix of edge weights with components
\begin{align}
	w_{xy} = \left\{
		\begin{array}{cc}
			\exp(-\beta |J(x) - J(y)|) & x \in \mathcal{N}(y) \\
			0 & \text{otherwise.}
		\end{array} \right.\label{eq_W}
\end{align}
Here, $x, y \in \Omega$, $J(x)$ is the image's intensity at $x$, $\mathcal{N}(y)$ is the set of voxels neighboring $y$, and $\beta \geq 0$ is a scalar parameter. Letting $D$ be the diagonal matrix with diagonal entries given by the row sums of $W$ (or column sums, since $W$ is symmetric), the \emph{Laplacian} matrix of the graph is defined as $L = D - W$. 

The RW algorithm calculates the label probabilities by solving, for each $k \in \{1, \dots, K\}$,
\begin{align}
	&\argmin_{\b{u}^k_n} \; {\b{u}^k}^\top L \b{u}^k + \gamma \left\| \b{u}^k - \b{p}^k \right\|^2 \; ,	\label{eq_rw_energy}
\end{align}
where $\gamma \geq 0$ is a scalar parameter. Writing $L$ and $P$ in block matrix notation similar to \eqref{eq_U_block},
\begin{align}
	L = \blockM{L_s}{B}{B^\top}{L_n} , \; P = \blockV{P_s}{P_n} , \; \b{p}^k = \blockV{\b{p}^k_s}{\b{p}^k_n} \label{eq_LP_block}
\end{align}
the globally optimal solution to \eqref{eq_rw_energy} for each $k$ can be found simultaneously by solving the linear system of equations:
\begin{align}
	(L_n + \gamma I_n) U_n = \gamma P_n - B^\top U_s \, \label{eq_rw_eq}
\end{align}
where $I_n$ is an appropriately sized identity matrix. This equation consists of $K$ linear systems, each of size $(N - S) \approx N$, and solving them can be very computationally expensive, which can disrupt the workflow when a user is waiting to provide additional input. However, the system matrix $(L_n + \gamma I_n)$ is mostly known offline; the only unknown is the $S \ll N$ rows and columns corresponding to the seeded voxels. This suggests precomputation could be performed on the matrix $(L + \gamma I)$ to quickly approximate the solution to \eqref{eq_rw_eq} online. %A solution is to move the computational burden offline, precomputing matrices that can be used to speed the calculation.

\subsection{Fast Random Walker Through Precomputation} \label{sec_precomp}

The precomputation scheme of Grady and Sinop \cite{grady2008fast} is based on an eigenvector decomposition of the Laplacian:
\begin{align}
	L = Q \Lambda Q^\top \label{eq_L_eigs} \;,
\end{align}
where the columns of $Q$ are the eigenvectors and $\Lambda$ is a diagonal matrix of the eigenvalues, ordered from smallest to largest. The Laplacian is known to be positive semi-definite, with one zero eigenvalue corresponding to the normalized constant vector $\b{g}$, with components $1 / \sqrt{n}$. \re{We now show how we use these eigenvectors to quickly find an approximate solution to \eqref{eq_rw_eq}.} Following Andrews et al. \cite{andrews2010fast}, we define $E$ as the pseudo-inverse of $(L + \gamma I)$, given by
\begin{align}
	E = Q (\Lambda + \gamma I)^{-1} Q^\top \label{eq_E_def} \;.
\end{align}
In block notation, 
\begin{align}
	E &= \blockM{E_s}{R}{R^\top}{E_n} \label{eq_E_block} \\
	&= \blockM{Q_s (\Lambda + \gamma I)^{-1} Q_s^\top}
						{Q_s (\Lambda + \gamma I)^{-1} Q_n^\top}
						{Q_n (\Lambda + \gamma I)^{-1} Q_s^\top}
						{Q_n (\Lambda + \gamma I)^{-1} Q_n^\top}\;.
\end{align}
In practice it is infeasible to calculate all $N$ eigenvectors of $L$, as $Q$ would have $\sim N^2$ non-zero elements, which would be too large of a matrix from both a computational and memory perspective, particularly for volumetric images. Thus, we only calculate the $m \ll N$ eigenvectors, corresponding to the $m$ smallest eigenvalues, since the smallest eigenvalues give the best approximation to $E$. 

We now define the unknown $F = (L + \gamma I) U$:
\begin{align}
	F &= \blockV{F_s}{F_n} \\
	F_s &= L_s U_s + \gamma U_s + B U_n \label{eq_Fs_1} \\
	F_n &= B^\top U_s + L_n U_n + \gamma U_n \\
	&= \gamma P_n \;, \; \text{from \eqref{eq_rw_eq}.}
\end{align}
Left multiplying by $E$ gives
\begin{align}
	EF = E (L + \gamma I) U &= U \label{eq_pseudo_E} \\
	R^\top F_s + \gamma E_n P_n &= U_n \label{eq_Fs_2} \\
	Q_n (\Lambda + \gamma I)^{-1} (Q_s^\top F_s + \gamma Q_n^\top P_n) &= U_n \;, \label{eq_U_lin_comb}
\end{align}
where \eqref{eq_Fs_2} takes only the non-seeded rows and uses the block notation for $E$ from \eqref{eq_E_block} and $P$ from \eqref{eq_LP_block}. Equation \eqref{eq_U_lin_comb} has two unknowns, $U_n$ and $F_s$; the strategy now is to remove $U_n$ and solve for $F_s$. To remove $U_n$, we right multiply \eqref{eq_Fs_2} by $B$ and subtract it from \eqref{eq_Fs_1}, giving
\begin{align}
	(I - B R^\top) F_s &= L_s U_s + \gamma (U_s + B E_n P_n) \label{eq_Fs} \;,
\end{align}
which has $F_s$ as the only unknown, and can thus be found by solving the systems of equations in \eqref{eq_Fs}. These systems are only of size $S \ll n$, and can be solved efficiently. Finally, we can solve for $U_n$ directly using \eqref{eq_Fs_2} instead of solving the system of equations in \eqref{eq_rw_eq}.  

We note that if $\gamma = 0$ (no prior probabilities), \eqref{eq_pseudo_E} must be altered, since $L$ has the constant vector $\b{g}$ as a zero eigenvector, so $EL = (I - \b{g}\b{g}^\top)$. A technique to address this case by was presented by Grady and Sinop \cite{grady2008fast}, which involves first calculating $\b{g}_n^\top U_n$, the column sums of $U_n$, and using those to calculte $F_s$.

The RW algorithm ran using this precomputation technique will be referred to as ``\fast RW'', and without as ``basic RW''. \fast RW has been shown to provide an excellent speed increase in both RWIS \cite{grady2008fast, andrews2010fast} and RWIR \cite{andrews2014fast} with only a minimal loss in accuracy. However, some questions still remain:
\begin{enumerate}
	\item How large should $m$ be? A larger $m$ leads to greater accuracy, but greater runtime both offline and online, since the online phase involves multiplications between matrices of size $n \times m$.
	\item What if the Laplacian is not known offline? Particularly, what if one wants a different value for $\beta$ in \eqref{eq_W}?
	\item What if the graph connectivity changes? In order to make RWIR more efficient, techniques have been developed that use sparser graph structures when solving \eqref{eq_rw_eq}, and this structure may be updated online \cite{tang2014reducing}.
\end{enumerate}
So far, justification of the above precomputation technique is largely algebraic. In order to answer these questions, we provide a more intuitive interpretation of the eigenvectors $Q$.

\subsection{Relationship to Normalized Cuts} \label{sec_ncuts}

Grady and Sinop \cite{grady2008fast} showed their precomputation technique could be thought of as incorporating user interaction into the \emph{normalized cuts} segmentation technique of Shi and Malik \cite{shi2000normalized}. Given a weighted image graph, a normalized cut seeks to find a partition of the vertices $\Omega$ into two sets, $\Omega_A$ and $(\Omega \setminus \Omega_A)$ that minimizes the normalized cut value
\begin{align}
	\text{Ncut}(\Omega_A) &= \frac{\cut(\Omega_A)}{\ass(\Omega_A)} + 
															 \frac{\cut(\Omega_A)}{\ass(\Omega \setminus \Omega_A)} \label{eq_n_cut}\\
  \cut(\Omega_A) &= \sum_{i \in \Omega_A, j \in \Omega \setminus \Omega_A} w_{ij} \\
	\ass(\Omega_A) &= \sum_{i \in \Omega_A, j \in \Omega} w_{ij} \; .
\end{align}

While finding the minimizing normalized cut is intractable, the relaxed solution (assigning real numbers instead of binary indicators to each vertex) is found by calculating the eigenvectors with smallest eigenvalues in the generalized eigenvector system (recall $L = D - W$):
\begin{align}
	L \b{v} = \lambda D \b{v} \;. \label{eq_norm_gen}
\end{align}
Defining $\hat{L} = D^{-1/2} L D^{-1/2}$ as the \emph{normalized} Laplacian, \eqref{eq_norm_gen} can be rewritten as a standard eigenvector system:
\begin{align}
	\hat{L}\hat{\b{v}} = \lambda \hat{\b{v}} \;, \label{eq_z_eig}
\end{align}
where $\hat{\b{v}} = D^{1/2} \b{v}$. Note that since the eigenvectors are orthogonal, they tend to represent complimentary cuts, representing different structures in the image (\figref{fig_eig_ex}). 

To complete the connection, we note that the RW algorithm can be formulated using the normalized Laplacian $\hat{L}$ instead of $L$ by defining $\hat{U}= D^{1/2}U$ and  $\hat{P} = D^{1/2}P$, substituting these new variables into the RW equation \eqref{eq_rw_eq}, solving for $\hat{U}$, and then recovering $U$. This results in the precomputation scheme from \secref{sec_precomp} calculating, in \eqref{eq_L_eigs}, the eigenvectors of $\hat{L}$ instead of $L$, which has been advocated because the normalized Laplacian tends to have a better behaved spectrum than the unnormalized Laplacian \cite{grady2008fast, chung1997spectral}. In this formulation, the columns of $Q$ correspond to relaxed normalized cuts of the image. In the subsequent sections, we assume the normalized Laplacian is used, but maintain the use of $U$, $P$, and $L$ for consistency, explicitly noting any differences that would arise when using the normalized instead of the unnormalized Laplacian.

Shi and Malik explore several techniques for using the eigenvectors to segment an image, such as thresholding the eigenvectors \cite{shi2000normalized}. In the subsequent section, however, we only require the intuition that the \fast RW techniques are approximating the original image with its most prominent structures as determined by normalized cuts, and, as seen in \eqref{eq_U_lin_comb}, the segmentations are linear combinations of the eigenvectors.

\section{Increasing Precomputation Robustness} \label{sec_online}

\subsection{Determining the Number of Eigenvectors} \label{sec_eigs}

Computing eigenvectors of the Laplacian is expensive, so there may be a limit to the number that can be computed offline, though this limit depends on factors such as the amount of computation power available and the throughput of images. More importantly, the online runtime scales linearly with $m$, the number of eigenvectors used, so with large enough $m$ \fast RW would be more computationally expensive than basic RW. Thus, we focus on choosing an appropriate number of eigenvectors online, and assume at least that many are computed offline.

The accuracy of \fast RW is characterized by how close the solution it produces (using \eqref{eq_Fs_2} and \eqref{eq_Fs}) is to the basic RW solution (produced by using \eqref{eq_rw_eq}), in terms of, for example, the Dice similarity coefficient for RWIS or the mean overlap between images for RWIR (described in more detail in \secref{sec_results}). However, the relationship between accuracy and $m$ for \emph{any} prior probabilities and user input seeds is difficult to characterize. For example, in RWIS, if the user input seeds happen to correspond well to the image structures represented by the first few eigenvectors, a very small $m$ should give good accuracy. However, if a user is trying to delineate more complex structures, a large $m$ may be required before these structures are represented well by the eigenvectors. Below we develop a strategy for efficiently determining the appropriate number of eigenvectors to use online, after user input seeds and prior probabilities become known. %For user input seeds, we focus on RWIS, and for prior probabilities, we focus on RWIR.

\subsubsection{User Seeds in RWIS} \label{sec_seeds}

From \eqref{eq_U_lin_comb}, each column of the segmentation $U_n = [\b{u}_n^1, \dots, \b{u}_n^K]$ is a linear combination of the columns of $Q_n$. Recall $U_s = [\b{u}_s^1, \dots, \b{u}_s^K]$ is an $S \times K$ matrix of $0$'s and $1$'s, with a single $1$ in each row, representing the probabilities of the seeded voxels. Intuitively, \re{since the unseeded voxels are derived from the seeds (see \eqref{eq_rw_eq})}, if no linear combination of the eigenvectors gives a segmentation for the seeded voxels similar to the labeling dictated by $U_s$, we conclude the segmentation specified by these seeds cannot be represented well by the current set of eigenvectors. 

To make this rigorous, if we define $\alpha_k \in \mathbb{R}^K$ such that $\b{u}_n^k = Q_n \alpha_k$, then we expect $\b{u}_s^k \approx Q_s \alpha_k$. We note the constraint
\begin{align}
	\| \b{u}^k \| = \| Q \alpha_k \| = \| \alpha \| \leq \sqrt{N} \;, 
\end{align}
since the components of $\| \b{u}^k \|$ are at most $1$. If the normalized Laplacian is being used, $\sqrt{n}$ should be replaced by $\sqrt{\b{1}^\top D \b{1}}$.  We now define the function $f(\b{u}_s^k)$ measuring how well the current eigenvectors can represent the seeds for label $k$:
\begin{align}
	f(\b{u}_s^k) &= \min_{\| \alpha \| \leq \sqrt{N}} \|Q_s \alpha - \b{u}_s^k \|^2 \label{eq_alpha_1}\;.
\end{align}
The minimization in \eqref{eq_alpha_1} is over a convex function with a convex constraint, and is of size $S \times m \ll N$,  so can be solved efficiently. \re{We calculate $f(\b{u}_s^k)$ for each $k$ to ensure each label is represented well by the eigenvectors,} and if any are above a threshold $f_{max}$, we add more eigenvectors and repeat the process. $f_{max}$ controls the trade-off between speed and accuracy, and is thus application dependent, though since $f(\b{u}_s^k) / S$ is roughly a measure of the expected squared error at each voxel, for a desired mean absolute error $\epsilon$, one should set $f_{max} = S \cdot \epsilon^2$. We use $\epsilon = 1/10$.

\subsubsection{Prior Probabilities in RWIR} \label{sec_priors}

The prior probabilities $P = [\b{p}^1, \dots, \b{p}^K]$ correspond to an unregularized probabilistic registration, so determining how well they can be represented by the eigenvectors is fairly straightforward. Since the columns of $Q$ are orthonormal, we can project each of the columns of $P$ onto their span and calculate the magnitude of the residual:
\begin{align}
	g(\b{p}_k) = \left\|\b{p}^k - Q Q^\top \b{p}^k \right\| ^ 2\;. \label{eq_g}
\end{align}
\rev{Since each label corresponds to a displacement vector, labels corresponding to similar displacement vectors often have similar probabilities, thus \eqref{eq_g} need not be evaluated for every $k$. To increase efficiency, we evaluate $g(\b{p}_k)$ for one out of every $r^d$ displacement labels, uniformly spaced, where $d$ is the image dimension. If any of them are above a threshold $g_{max}$, we add more eigenvectors. In this work, we use $r=4$, so \eqref{eq_g} is evaluated for $(K / 64)$ labels in volumetric registration.} Similar to $f_{max}$, $g_{max}$ controls the trade-off between speed and accuracy, and $\sqrt{g_{max} / n}$ is roughly proportional to the voxel-wise error in the probabilities. For a desired mean absolute error $\epsilon$, one should set $g_{max} = n \cdot \epsilon^2$. We use $\epsilon = 1/10$. Note that we save the residual $(\b{p}^k - Q Q^\top \b{p}^k)$, so we are able to update \eqref{eq_g} efficiently when more eigenvectors are added by projecting the residual onto the new eigenvectors.

Calculating $g(\b{p}_k)$ can be relatively expensive, as it requires multiplications between matrices of length $n$. In segmentation, the priors are sometimes available offline (e.g. expected range of HU units in CT for different tissue types); if the priors are derived from the user input seeds, calculating $g(\b{p}_k)$ may not be worth the time. In registration, calculating the priors often constitutes a significant overhead, so calculating $g(\b{p}_k)$ may be relatively insignificant.

\subsection{Updating the Laplacian} \label{sec_beta}

The RW algorithm can be significantly affected by how the edge weights are calculated in \eqref{eq_W}, but the Laplacian must be fixed when precomputation is performed. Previously, if a user wanted to change the edge weights online, they had to abandon using \fast RW. In this section, we propose a technique to update the precomputed data when the edge weights are changed. Specifically, we focus on changes in the key parameter $\beta$.

One potential technique is to precompute eigenvectors for multiple Laplacians with different edge weights, though there are  limitations to this technique: there's no guarantee the desired Laplacian (i.e. for a specific $\beta$) will be precomputed offline, and the offline computational burden could limit the number of Laplacians that can be used. Instead, we make the observation that when $\beta$ is increased or decreased, the relative ordering of the edge weights doesn't change, \re{so small relaxed normalized cuts for one value of $\beta$ should also be small for other values of $\beta$}. Thus, while the eigenvectors of the Laplacian are different for different values of $\beta$, they should correspond roughly to the same prominent image structures. This suggests we may be able to compute the eigenvectors for a certain $\beta'$ and then re-use them for other $\beta$'s. The key is updating the eigen\emph{values}.

%In normalized cuts, the eigenvector system \eqref{eq_z_eig} that is used to solve for the relaxed cuts is derived from the Rayleigh quotient minimization problem:
\re{The relaxed normalized cuts function in \eqref{eq_n_cut} is equivalent to the Rayleigh quotient:}
\begin{align}
	\Ncut(\b{v}) = \frac{\b{v}^\top \hat{L} \b{v}}{\b{v}^\top \b{v}} \label{eq_z_ray} \;.
\end{align}
If $\b{v}$ is an eigenvector of $\hat{L}$ with corresponding eigenvalue $\lambda$, $\Ncut(\b{v}) = \lambda$. Thus, the values on the diagonal of $\Lambda$ are not only the eigenvalues corresponding to the eigenvectors $Q$, but also their normalized cut values. 

When $\beta$ is updated online, we construct the new normalized Laplacian $\hat{L}$, evaluate the new $\Ncut(\cdot)$ function (calculated using the offline Laplacian) at each column of $Q$, and replace the values on the diagonal of $\Lambda$ with the new normalized cut values. \rev{That is, for $Q = [\b{q}^1, \dots, \b{q}^m]$, we define $\hat{\Lambda}$ as an $m \times m$ diagonal matrix with elements $\{\hat{\lambda}_1, \dots, \hat{\lambda}_m\}$ given by
\begin{align}
	\hat{\lambda}_i = {\b{q}^i }^\top \hat{L} \b{q}^i \;.
\end{align}
$\hat{E}$, the pseudo-inverse of $\hat{L}$, is then approximated by
\begin{align}
	E = Q (\hat{\Lambda} + \gamma I)^{-1} Q^\top \;,
\end{align}
instead of using the actual eigenvectors of $\hat{L}$, as in \eqref{eq_E_def}.
This weights each column of $Q$ by how well it ``cuts'' the updated image graph, so columns that no longer correspond to good cuts will be largely ignored. While $Q$ and $\hat{\Lambda}$} can no longer be considered eigenvectors and eigenvalues, their use in the fast RW algorithm does not otherwise change. While this approximation \rev{is expected to} get worse for $\beta$ further from $\beta'$, we mitigate this by precomputing eigenvectors for several different $\beta$'s and using \eqref{eq_z_ray} to ``interpolate'' between them.

\subsection{Aggregated Image Graphs} \label{sec_agg}

In RWIR, as with other discrete registration techniques, there has been recent work targeted at reducing the number of vertices in the image graph, and thus the computational cost of RWIR, by aggregating vertices into ``super-vertices''. While this aggregation has often been done in an image agnostic way, using grids of predefined resolutions, recent image dependent online techniques have proven successful \cite{popuri2013variational, tang2014reducing, parisot2014concurrent}. \re{In general, vertex aggregation is performed by defining a new set of vertices $\bar{\Omega}$, $|\bar{\Omega}| = \bar{N} < N$ and a ``projection function'' $\eta(x, y)$, where $x \in \Omega$ and $y \in \bar{\Omega}$,  encoding the influence of each super-vertex on vertex $x$ when propagating values from the sparser graph to the denser graph. The corresponding ``aggregation function'' is defined as $\bar{\eta}(x, y)$, encoding the influence of each vertex on super-vertex $y$ when propagating values from the denser graph to the sparser graph. We require these functions be normalized:}
\begin{align}
	\sum_{y \in \bar{\Omega}} \eta(x, y) = 1, \quad \sum_{x \in \Omega} \bar{\eta}(x, y) = 1 \;.
\end{align}
Once RW is solved on the aggregated graph, the solution is propagated back to the original vertices. Denoting probabilities assigned to $y \in \hat{\Omega}$ as $\bar{U}_y$, the corresponding probabilities for vertex $x$ are given by
\begin{align}
	U_x = \sum_{y \in \bar{\Omega}} \eta(x, y) \bar{U}_y \;. 
\end{align}
Note that for a given $x$, often $\eta(x, y) = 0$ for most $y$, with vertices only dependent on a few nearby super-vertices.

Vertex aggregation performed online invalidates the eigenvectors precomputed from the original graph. We cannot apply the technique from \secref{sec_beta} directly, since the number of vertices have changed, so the normalized cut values cannot be computed for the columns of $Q$. We thus develop a technique for aggregating the columns of $Q$ to the super-vertices. 

Let $\b{q} = [q_1, \dots, q_n]^\top$ be one of the columns of $Q$, and let $\bar{\b{q}} = [\bar{q}_1, \dots, \bar{q}_{\bar{n}}]^\top$ be the (undetermined) aggregation of $\b{q}$ to the super-vertices. The most straightforward technique for calculating $\bar{\b{q}}$ would be to use $\bar{\eta}(\cdot, \cdot)$ directly:
\begin{align}
	\bar{\b{q}}_y = \sum_{x \in \Omega} \bar{\eta}(x, y) \b{q}_x  \;. \label{eq_q_agg_1}
\end{align}
However, using \eqref{eq_q_agg_1} may not respect the property that $\bar{\b{q}}$ should have a small normalized cut value on the aggregate graph. A super-vertex $j$ may ``consume'' some edges, if the vertices on both ends of the edge are assigned exclusively to $j$. The influence of consumed edges should not be considered when calculating $\bar{\b{q}}$, as they have no effect on the normalized cut of the aggregate graph. Consider a vertex $i$ with all of its edges consumed; $q_i$ was based exclusively on those edges, and thus should not be considered when performing aggregation.

To account for this, instead of weighting components of $\b{q}$ just by $\bar{\eta}(\cdot, \cdot)$, we also weight them by the local change in $\eta(\cdot, \cdot)$, since the edges between vertices assigned to different super-vertices will still affect the normalized cut value:
\begin{align}
			\Delta(x) &= \sum_{x' \in \mathcal{N}(x)} \sum_{y \in \bar{\Omega}} |\eta(x, y) - \eta(x', y)| \\ 
			\bar{\b{q}}_y  &= \sum_{x \in \Omega} \Delta(x) \; \bar{\eta}(x, y) \; \b{q}_x \;. \label{eq_q_agg_2}
\end{align}
We note that the vectors generated by \eqref{eq_q_agg_2} will not form an orthonormal set, but are made so using the Gram-Schmidt process. Combining the orthonormal vectors into a matrix $\bar{Q}$ and then calculating their corresponding $\Ncut$ values allows \fast RW to be run on the aggregate graph with minimal overhead and loss in accuracy.

\rev{We focus on \fast RWIR for graph aggregation as we found the aggregation step was typically too expensive compared to the \fast RWIS run-time, along with the possibility of the aggregated super-voxels crossing weak image boundaries and reducing segmentation accuracy.}

\begin{figure}
	\centering
	\subfloat[Manual Seg.]{
		\begin{minipage}{0.29\linewidth}
			\includegraphics[width=1.0\textwidth]{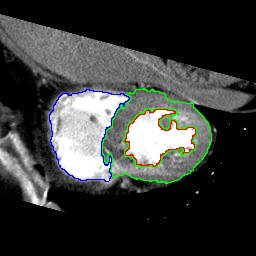} \\			
			
			\includegraphics[width=1.0\textwidth]{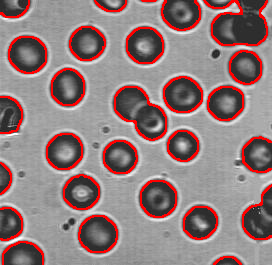}\\			
			
			\includegraphics[width=1.0\textwidth]{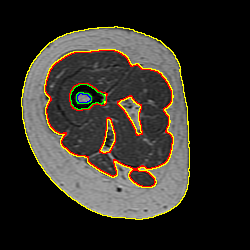}
		\end{minipage}
	}
	\subfloat[Basic RWIS]{
		\begin{minipage}{0.29\linewidth}
			\includegraphics[width=1.0\textwidth]{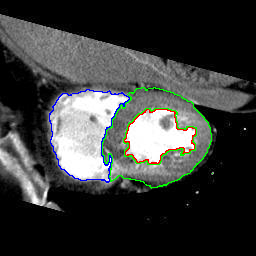} \\			
			
			\includegraphics[width=1.0\textwidth]{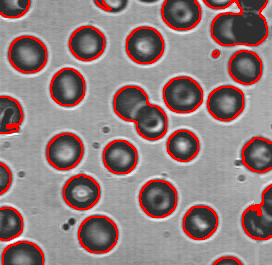}\\			
			
			\includegraphics[width=1.0\textwidth]{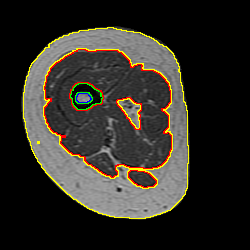}
		\end{minipage}
	}
	\subfloat[\fast RWIS]{
		\begin{minipage}{0.29\linewidth}
			\includegraphics[width=1.0\textwidth]{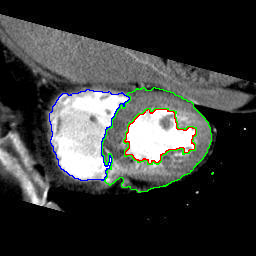} \\			
			
			\includegraphics[width=1.0\textwidth]{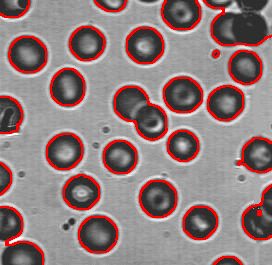}\\			
			
			\includegraphics[width=1.0\textwidth]{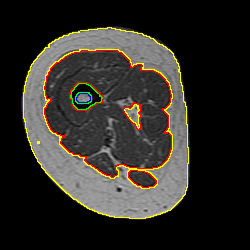}
		\end{minipage}
	}	
	
	\caption{
		(Color figure) An example of the segmentations achieved by basic RWIS and \fast RWIS with eigenvectors adapted to the seeds. Regions are outlined in color. \fast RWIS takes only about $20\%$ of time as basic RWIS (see \figref{fig_adapt_eigs_seeds}).
		%Using basic RWIS provides only a minimal increase in average DSC of (from left to right) $0.003$, $0.011$, and $0.013$, whereas 
	}
	\label{fig_adapt_eigs_segs}
\end{figure}

\section{Results} \label{sec_results}

In this section, we test the effectiveness of the techniques presented in \secref{sec_online}. Since \fast RW is an approximation to the basic RW method, our evaluations focus on measuring the loss in accuracy compared to the increase in run-time when using  \rev{\fast RW (with and without the adaptive techniques presented here) instead of basic RW}. Ideally, only a small fraction of the overall accuracy will be lost when using \fast RW. Note that ``accuracy'' will be defined by comparing to manual segmentation results using the Dice similarity coefficient (DSC) in segmentation and mean overlap (MO) in registration \cite{klein2009evaluation}. For two images segmented into $K$ foreground regions given by $\Omega_k^1$ and $\Omega_k^2, k \in \{1, \dots, k\}$,
\begin{align}
	\text{MO} = 2\frac{\sum_{k=1}^K |\Omega_k^1 \cap \Omega_k^2|}{\sum_{k=1}^K |\Omega_k^1| + |\Omega_k^2|} \;.
\end{align}  
\re{Probabilistic segmentations converted to non-probabilistic segmentations by thresholding and probabilistic registrations are converted to non-probabilistic registrations by taking the expected displacement at each voxel.}

%
%Since RWIS and RWIR use the same formulations, and more specifically the same definition of the Laplacian, we focus on evaluating RWIS accuracy where possible, as it is a more computationally efficient and allows more experiments to be run. However, we focus on RWIR when evaluating our techniques presented from \secref{sec_priors} and \secref{sec_agg}, as they are useful mainly for registration.

\subsection{Setup} \label{sec_data}

\begin{figure}
	\centering
	\includegraphics[width=0.48\textwidth]{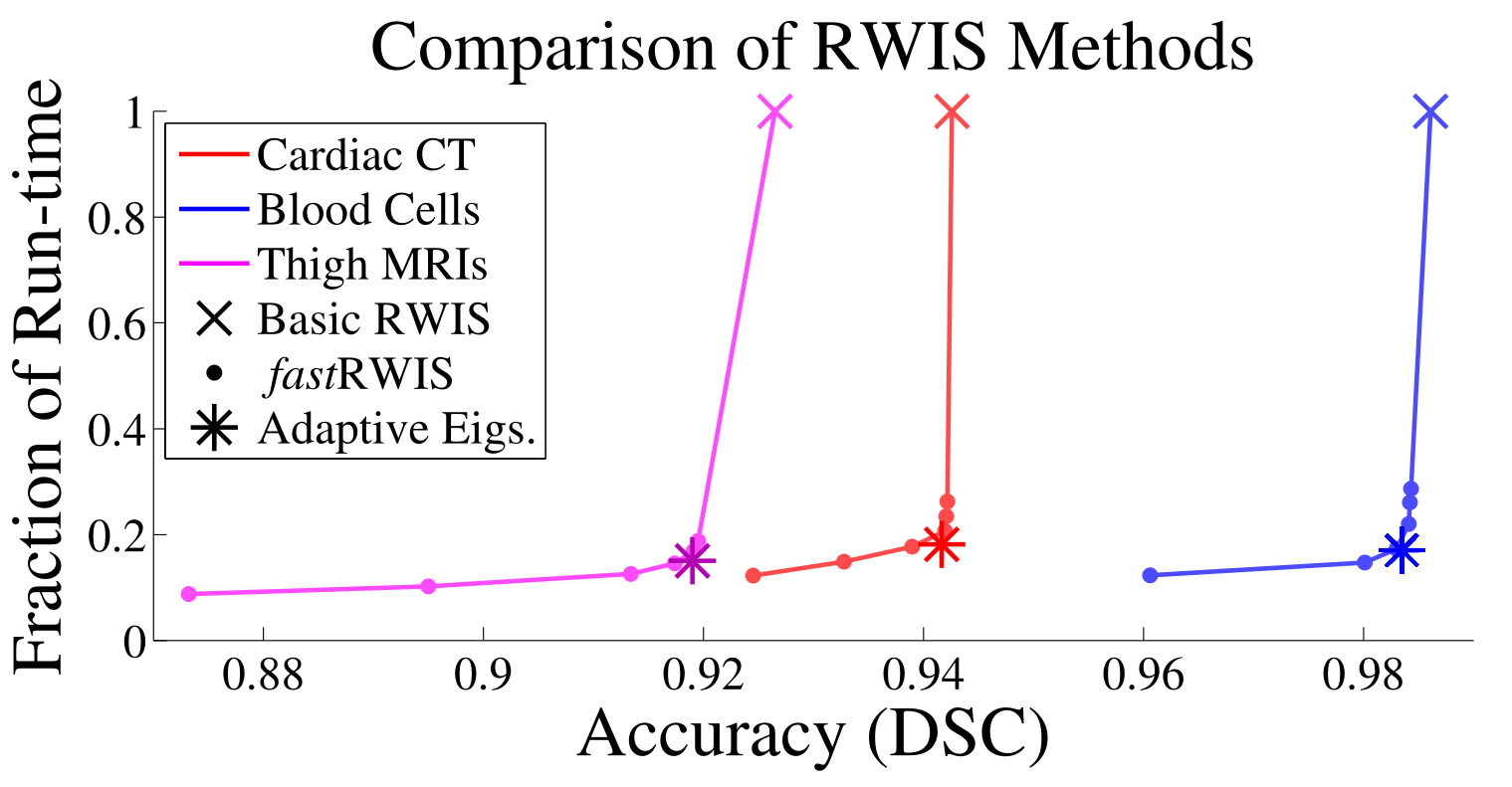}
	\caption{(Color figure) A comparison of run-time \rev{(as a fraction of basic RWIS)} vs. accuracy for basic RWIS, \fast RWIS with fixed numbers of eigenvectors, and \fast RWIS with eigenvectors adapted automatically to the seeds. Our adaptive approach achieves excellent accuracy and run-time while automatically choosing the number of eigenvectors to use. See \tabref{table_acc_list} for more details.}
	\label{fig_adapt_eigs_seeds}
\end{figure}

We use two $2$D images and $2$ data sets of $40$ volumetric images each. The $2$D images consist of a  $256 \times 256$ CT cardiac image slice, manually segmented into $4$ regions, and a $265 \times 272$ blood cell image, manually segmented into $2$ regions (blood cell and background). $2$D data allows a large number of tests with different seed locations. The cardiac image slice  is used because it provides a multi-label segmentation and the blood cell image is used because it requires prior probabilities calculated from the user seeds to segment every cell (\figref{fig_adapt_eigs_segs}). 

\begin{figure}
	\centering
	\subfloat[Original Alignment]{
		\begin{minipage}{0.29\linewidth}
			\includegraphics[width=1.0\textwidth]{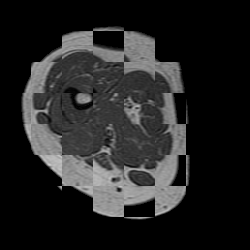} \\
			
			\includegraphics[width=1.0\textwidth]{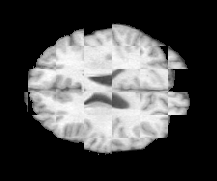}
		\end{minipage}	}	\;
	\subfloat[Basic RWIR]{
		\begin{minipage}{0.29\linewidth}
			\includegraphics[width=1.0\textwidth]{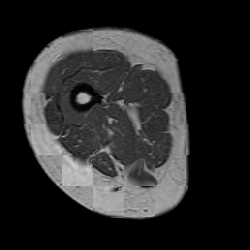} \\
			
			\includegraphics[width=1.0\textwidth]{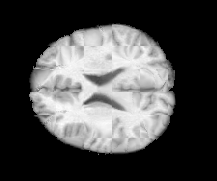}
		\end{minipage}	}	\;
	\subfloat[\fast RWIR]{
		\begin{minipage}{0.29\linewidth}
			\includegraphics[width=1.0\textwidth]{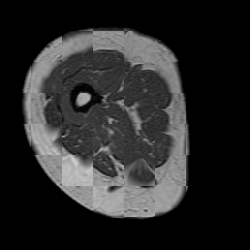} \\
			
			\includegraphics[width=1.0\textwidth]{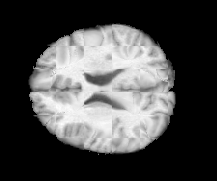}
		\end{minipage}	}
	\caption{
		(Color figure) An example of the registrations achieved by basic RWIR and \fast RWIR with eigenvectors adapted to the priors. \fast RWIR takes only about $30\%$ of time as basic RWIR (see \figref{fig_adapt_eigs_priors}).
	}
	\label{fig_adapt_eigs_regs}
\end{figure}

The first volumetric data set consists of $40$ $(250 \times 250 \times 40)$ T1-MR images of thighs, manually segmented into $16$ regions. For segmentation, we combine some regions, resulting in a $5$-label segmentation (muscle, fat, cortical bone, bone marrow, and background, see \figref{fig_adapt_eigs_segs}, \figref{fig_adapt_eigs_regs}). %We combine regions because the manual segmentation separates individual thigh muscles, but the intensity and texture profiles of each muscle is very similar, and intermuscular boundaries are often very hard to detect, so segmenting them with RWIS alone is very difficult.

\begin{figure}
	\centering
	\includegraphics[width=0.48\textwidth]{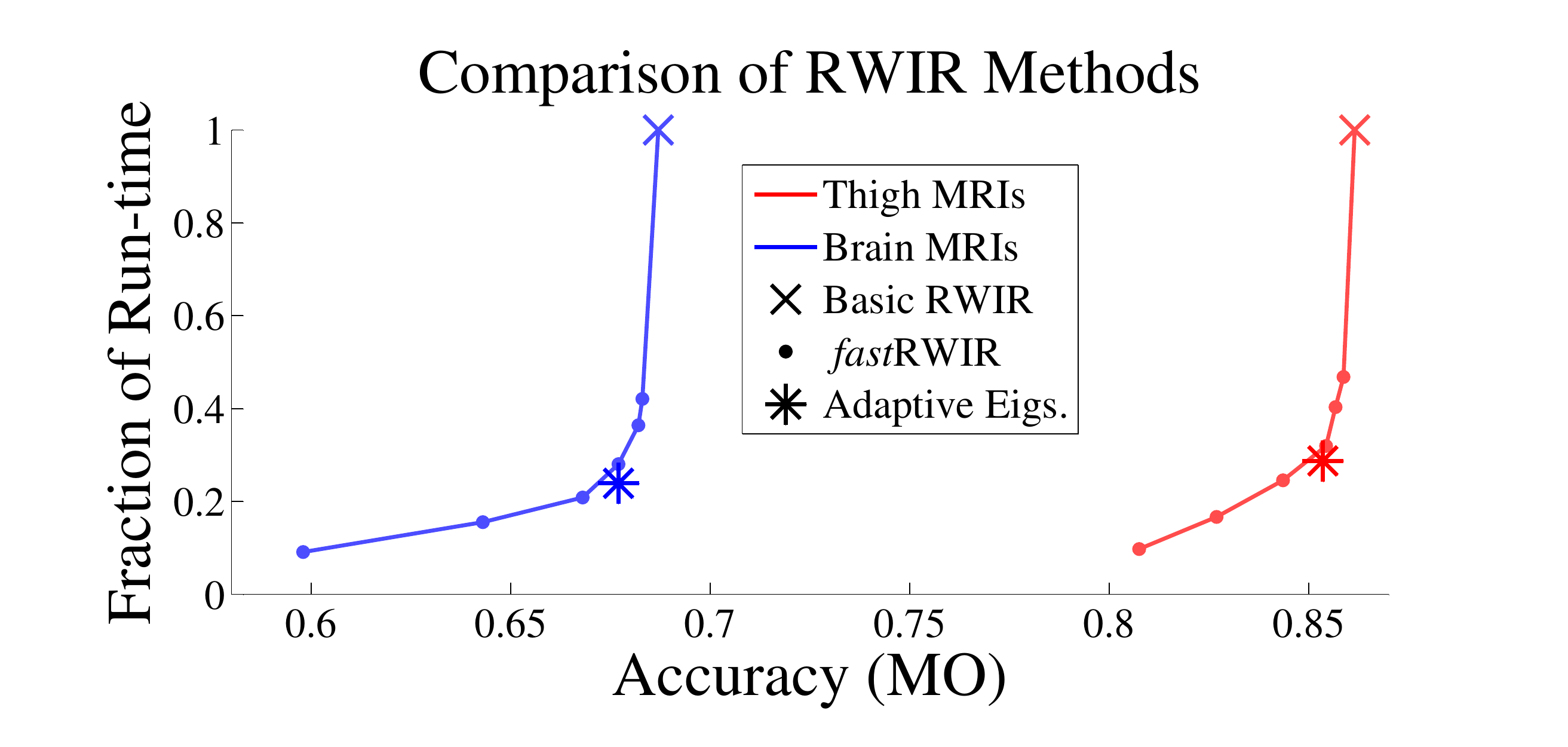}
	\caption{(Color figure) A comparison of run-time \rev{(as a fraction of basic RWIR)} vs. accuracy for basic RWIR, \fast RWIR with fixed numbers of eigenvectors, and \fast RWIR with eigenvectors automatically adapted to the prior probabilities. Our adaptive approach achieves excellent accuracy and run-times while automatically choosing the number of eigenvectors to use. See \tabref{table_acc_list} for more details. }
	\label{fig_adapt_eigs_priors}
\end{figure}

The second volumetric data set consists of $40$ $(181 \times 181 \times 217)$ T1-MR brain images from the LONI dataset \cite{klein2009evaluation}, manually segmented into $56$ regions, skull stripped, and with their intensity histograms normalized (see \figref{fig_adapt_eigs_regs}). 

When testing segmentation, we use the $2$D images and the thigh images, with seeds generated automatically using the manual segmentations. For the cardiac image slice, we randomly place $10$ seeds in each region, and with the blood cell image, we randomly place $10$ seeds inside a single blood cell and $10$ seeds nearby in the background. We generate $100$ different sets of seeds for each image, with the single blood cell chosen randomly each time. For the thigh images, we randomly place $40$ seeds each in the muscle, fat, and background regions, and $20$ seeds each in the two bone regions. We generate $5$ sets of seeds each for the $40$ images. All of the segmentation results in this section are averaged over each set of seeds. \rev{Different techniques for sampling seeds would lead to different segmentation accuracies, but as our goal is to compare different segmentation methods, it is sufficient to ensure they use the same sets of seeds.} 
Unless otherwise stated, we use $\beta = 50$, $\gamma = 0.001$ for RWIS on cardiac and thigh images, and $\gamma = 0.01$ for RWIS on blood cell images. The prior probabilities $P$ are calculated by fitting a Gaussian to the intensity values of the seeds for each region. 

When testing registration, we use the $3$D thigh and LONI brain images, registering each pair of images together for $39 \times 40$ tests total. The local similarity between two voxels is evaluated using the sum of absolute intensity differences in a patch of size $5 \times 5 \times 5$. \rev{We use $K \approx 12,000$  $(31 \times 31 \times 13)$ displacement labels for the thigh images and $K \approx 4000$ $(21 \times 21 \times 9)$ for the brain images.}
displacement labels for  The prior probabilities $P$ for displacement vectors are calculated as the negative exponential of the squared local patch difference, normalized to sum to $1$ at each voxel. We use $\beta = 50$, $\gamma = 1$ for RWIR.

Experiments use unoptimized MATLAB code run on a machine with $2$ Quad Core Intel Xeon $2.33$ GHz CPUs.

\subsection{Determining the Number of Eigenvectors} \label{sec_results_eigs}

\subsubsection{User Seeds} \label{sec_results_seeds}

First, we evaluate our technique from \secref{sec_seeds} for choosing the number of eigenvectors $m$ based on the seed locations (see \eqref{eq_alpha_1}). \rev{Our goal is to show we can adaptively and automatically choose $m$ when segmenting real medical images and achieve results comparable to the results for the best fixed $m$.} For the $2$D images, we precompute $160$ eigenvectors, and run RWIS and \fast RWIS using $m \in \{60, 80, \dots, 160\}$. For the $3$D thigh images, we precompute $800$ eigenvectors, and run RWIS and \fast RWIS using $m \in \{300, 400, \dots, 800\}$.  We then run \fast RWIS with $m$ chosen using our adaptive technique for each image. The accuracy and run-time of each segmentation algorithm is seen in \figref{fig_adapt_eigs_seeds}, with example segmentations shown in \figref{fig_adapt_eigs_segs}. 	

We see that our technique for choosing $m$ online provides accuracy comparable to using all the eigenvectors while running, on average, significantly faster. We emphasize that while each image seems to have an ``optimal'' number of eigenvectors which gives similar results to our technique in terms of run-time and accuracy, this number is not known ahead of time and is different for each image class ($\sim 120$ for cardiac, $\sim 100$ for blood cells, and $\sim 700$ for thigh). Our technique requires no prior knowledge to achieve this accuracy, and further, provides slightly lower run-time than any fixed number of eigenvectors, due to cases where fewer eigenvectors are used. \rev{Clearly the performance of \fast RWIS is highly dependent on the number of eigenvectors, yet choosing an appropriate number online would be guesswork without our technique.}

\begin{figure}
	\centering
	\includegraphics[width=0.48\textwidth]{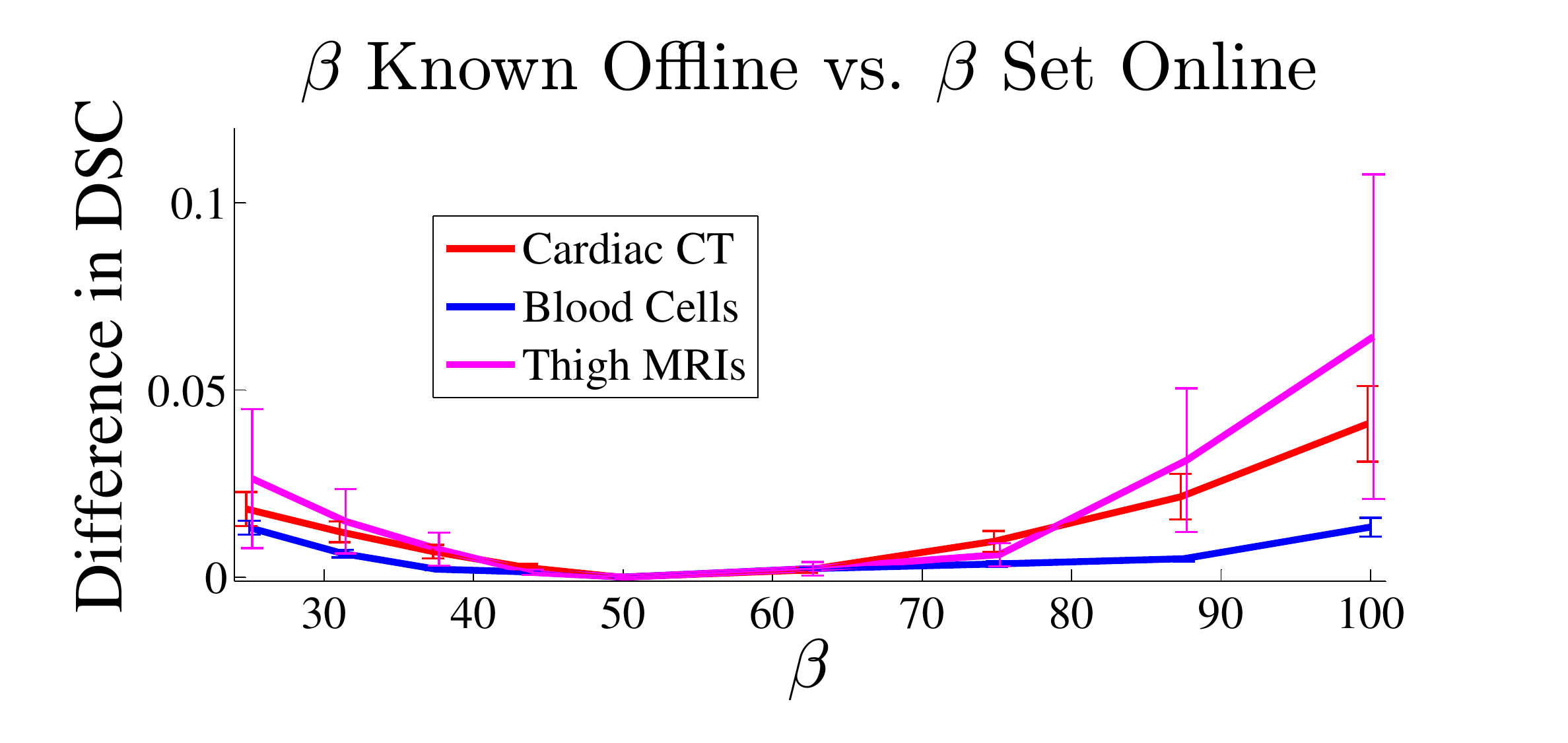}
	\caption{(Color figure) 
		A comparison of \fast RWIS accuracy when $\beta$ is known offline versus when $\beta$ is changed online and $\Lambda$ is updated using our technique from \secref{sec_beta}. The eigenvectors from the Laplacian with $\beta' = 50$ are used as the base eigenvectors when $\beta$ is not known. The difference in DSC \rev{(with respect to the optimal segmentation)} between the results of the two methods is reported. As expected, the DSC is higher when the Laplacian is known offline, though only slightly higher for a large range of values, particularly for the blood cell image. This indicates our technique allows $\beta$ to be adjusted online by at least a factor of $2$ while sacrificing only minimal accuracy ($< 3\%$ for the blood cell image). }
	\label{fig_adapt_lapl_beta}
\end{figure}

\subsubsection{Prior Probabilities} \label{sec_result_priors}

Second, we evaluate our technique from \secref{sec_priors} for choosing the number of eigenvectors $m$ based on the prior probabilities (see \eqref{eq_g}). \rev{Again, our goal is to show we can adaptively choose $m$ when registering a pair of and achieve results comparable to the best fixed $m$.} For each pair of images, we choose one randomly as the moving image and precompute $1500$ eigenvectors. We run RWIR, \fast RWIR with $m \in \{250, 500, \dots, 1500\}$, and \fast RWIR with $m$ chosen using our adaptive technique. The accuracy and the run-time of each registration algorithm is seen in \figref{fig_adapt_eigs_priors}, with example registrations show in \figref{fig_adapt_eigs_regs}. Similar to segmentation, our technique allows \fast RWIR to achieve an average accuracy and run-time comparable to the best fixed number of eigenvectors, but without any prior knowledge.

\begin{table}[h]
	\centering 
	
		\begin{tabular}{| c | r | r @{\;} l | r @{\;} l |} \hline 
			Data & \multicolumn{1}{c|}{Algorithm} & \multicolumn{2}{c|}{Accuracy} & \multicolumn{2}{c|}{Run-Time (sec)} \\ \hline
			\multirow{3}{*}{Cardiac} & RWIS & $0.943$ & $\pm \; 0.027$ & $0.983$ & $\pm \; 0.012$ \\ \cline{2-6}  
												 & \fast RWIS & $0.942$ & $\pm \; 0.031$ & $0.203$ & $\pm \; 0.003$  \\ \cline{2-6} 
							  & Adaptive \fast RWIS & $0.942$ & $\pm \; 0.034$ & $0.178$ & $\pm \; 0.006$ \\ \hline
			\multirow{3}{*}{Blood Cell} & RWIS & $0.986$ & $\pm \; 0.012$ & $1.07$ & $\pm \; 0.014$ \\ \cline{2-6}  
			                      & \fast RWIS & $0.984$ & $\pm \; 0.015$ & $0.24$ & $\pm \; 0.004$  \\ \cline{2-6} 
			             & Adaptive \fast RWIS & $0.983$ & $\pm \; 0.016$ & $0.19$ & $\pm \; 0.007$ \\ \hline
			\multirow{7}{*}{Thigh} & RWIS & $0.927$ & $\pm \; 0.034$ & $289.2$ & $\pm \; 10.3$ \\ \cline{2-6}  
			                 & \fast RWIS & $0.917$ & $\pm \; 0.041$ & $42.0$ & $\pm \; 2.3$  \\ \cline{2-6} 
			        & Adaptive \fast RWIS & $0.919$ & $\pm \; 0.040$ & $43.6$ & $\pm \; 3.4$ \\ \cline{2-6} 
							& \multicolumn{5}{c|}{} \\ \cline{2-6}
			                       & RWIR & $0.862$ & $\pm \; 0.044$ & $864$ & $\pm \; 44.5$ \\ \cline{2-6}  
			                 & \fast RWIR & $0.854$ & $\pm \; 0.053$ & $276$ & $\pm \; 18.3$  \\ \cline{2-6} 
			        & Adaptive \fast RWIR & $0.854$ & $\pm \; 0.054$ & $248$ & $\pm \; 21.3$ \\ \hline
			\multirow{3}{*}{Brain} & RWIR & $0.687$ & $\pm \; 0.091$ & $2112$ & $\pm \; 130.0$ \\ \cline{2-6}  
			                      & \fast RWIR & $0.677$ & $\pm \; 0.103$ & $592$ & $\pm \; 35.3$  \\ \cline{2-6} 
			             & Adaptive \fast RWIR & $0.677$ & $\pm \; 0.105$ & $504$ & $\pm \; 41.4$ \\ \hline
			\multicolumn{6}{c}{}
		\end{tabular}				
		
\caption{
	A summary of the results from \secref{sec_results_eigs}. Adaptive \fast RWIS and adaptive \fast RWIR choose the number of eigenvectors to use online based on the user seeds or prior probabilities. The results reported for \fast RWIS and \fast RWIR are for a fixed number of eigenvectors $m$, chosen separately for each image to give the run-time and accuracy closest to their adaptive counter-parts. Accuracy is reported using DSC or MO for RWIS and RWIR, respectively, \rev{and aggregation time is included.}
} 
\label{table_acc_list}
\end{table}

\subsection{Updating the Laplacian}

We evaluate our technique from \secref{sec_beta} (updating the precomputed values when the Laplacian edge weights are changed) by precomputing eigenvectors/values for an offline Laplacian constructed with $\beta' = 50$, then running \fast RW using an online Laplacian with $\beta \in [25, 100]$ and a $\Lambda$ updated using \eqref{eq_z_ray}.  We note that doubling $\beta$ corresponds to squaring the weights of all the edges (see \eqref{eq_W}). As a baseline, we compare the results to \fast RWIS run using eigenvector/values calculated from the online Laplacian directly. We precompute $320$ eigenvectors for the $2$D images and $1600$ for the thigh images. The accuracy of each technique is shown in \figref{fig_adapt_lapl_beta}. We see that when $\beta$ is \emph{not} known offline, our technique for updating $\Lambda$ achieves very similar accuracy compared to when $\beta$ \emph{is} known offline.

\subsection{Aggregated Image Graphs}

\begin{figure}
	\centering
	\subfloat[Images]{
		\begin{minipage}{0.45\linewidth}
			\includegraphics[width=1.0\textwidth]{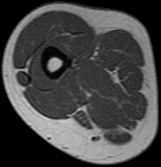} \\
			
			\includegraphics[width=1.0\textwidth]{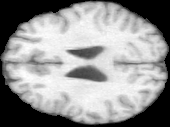}
		\end{minipage}
	} \;
	\subfloat[Aggregated Graphs]{
		\begin{minipage}{0.45\linewidth}
			\includegraphics[width=1.0\textwidth]{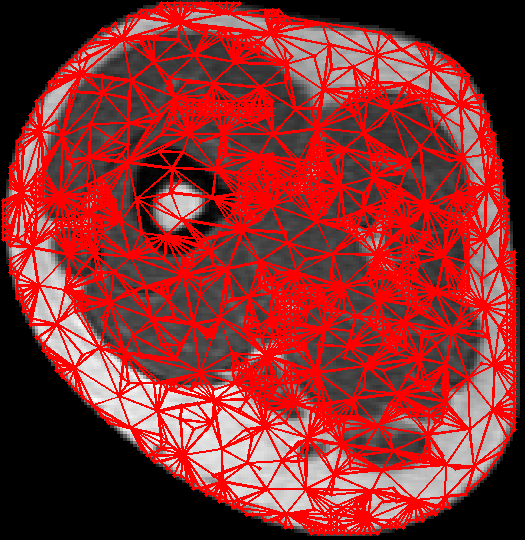} \\
			
			\includegraphics[width=1.0\textwidth]{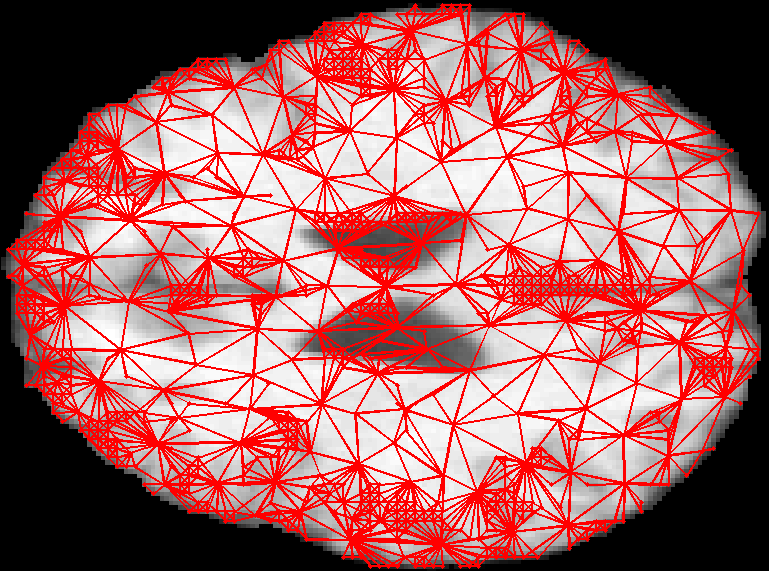}
		\end{minipage}
	}
	\caption{
		(Color figure) An example of aggregated graphs used to improve RWIR efficiency. Nearby nodes with similar prior probabilities are aggregated into super-vertices. The aggregated graphs have on average $18\%$ and $15\%$ as many vertices as the original graph for the thigh and brain images, respectively.
	}
	\label{fig_agg_graphs}
\end{figure}

\begin{figure}
	\centering
	\includegraphics[width=0.48\textwidth]{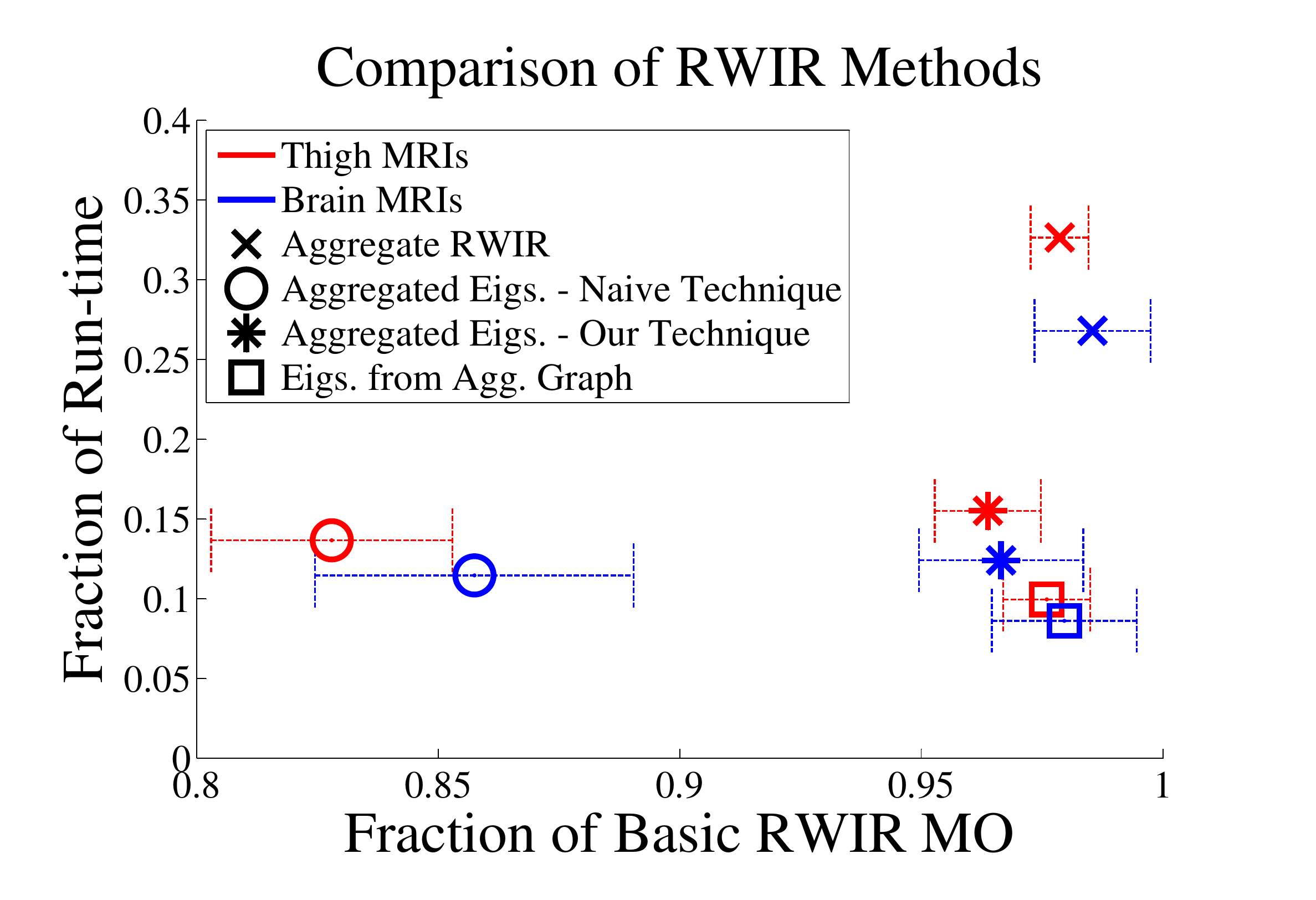}
	\caption{(Color figure) 
		A comparison of RWIR techniques used on graphs with aggregated super-vertices. Accuracy and run-time are given in proportion to RWIR run on the full graph (see \figref{fig_adapt_eigs_priors} for absolute values). When run on the aggregated graph, RWIR achieves a significant speed-up with a minimal loss in accuracy, with further speed-up achieved by \fast RWIR. The run times and MO values are given as a fraction of those achieved by the basic RWIS algorithm (see \figref{fig_adapt_eigs_priors} for absolute values). We note that using the eigenvectors from the aggregated graph (``$\square$'') is not actually possible, since the aggregated graph is not known off-line, and is only included as a baseline.
	}
	\label{fig_adapt_agg_graph}
\end{figure}

In this section, we evaluate our technique for aggregating the eigenvectors into super-vertices, given by \eqref{eq_q_agg_2} in \secref{sec_agg}. We use $m = 2500$ eigenvectors for \fast RWIR. We aggregate vertices based on their prior probabilities $P$ and spatial proximity \cite{tang2014reducing} (see \figref{fig_agg_graphs}), and thus the aggregation must be performed online, after the precomputation. For each pair of thigh and brain images, we run $4$ registration schemes: basic RWIR, \fast RWIR with the eigenvectors aggregated using the na\"{\i}ve technique in \eqref{eq_q_agg_1}, \fast RWIR with the eigenvectors aggregated using our proposed technique in \eqref{eq_q_agg_2}, and using eigenvectors calculated from the Laplacian of the aggregated graph directly (used only as a baseline, since the aggregated graph is not known offline). The accuracy and run-time of each technique is seen in \figref{fig_adapt_agg_graph}. Our proposed technique performs significantly better than the na\"{\i}ve technique, and only slightly worse than the eigenvectors calculated directly from the aggregate graph.

%\begin{table*}[h]
	%\centering 
		%\begin{tabular}{| c || r@{\;}l | r@{\;}l | r@{\;}l | r@{\;}l | r@{\;}l | r@{\;}l |} \hline 
			%Image & \multicolumn{2}{c|}{RWIS} & \multicolumn{10}{c|}{\fastrwir}  \\ \hline	
			%Type & & & \multicolumn{2}{c|}{$m = $} & \multicolumn{2}{c|}{$m = $} & \multicolumn{2}{c|}{$m = $} & \multicolumn{2}{c|}{$m = $} & \multicolumn{2}{c|}{Our Technique} \\ \hline
			%
				%%& $1$ & $0.63$ & $\pm \; 0.15$ & $0.7$ & $\pm \; 0.15$ & $ \b{0}$ & $\b{.0019}$\\ \cline{2-8}   
		%\end{tabular}				 
%\caption{
%
%} 
%\label{table_rwis_seeds}
%\end{table*}

\section{\rev{Discussion and Conclusion}} \label{sec_conclusion}

The main drawback of \fast RW algorithms is that certain algorithmic decisions must be made offline, when there is incomplete information. We have proposed multiple extensions that significantly improve the online flexibility of \fast RW algorithms, while only incurring a minor overhead in run-time and implementation complexity.  Since the \fast RW paradigm already achieves results very similar to the popular RWIS and RWIR algorithms, removing its limitations constitutes a major improvement to RW algorithms in general. Further, while the techniques presented here were only demonstrated individually, there is nothing to preclude combining them to further increase the robustness of the online algorithm to the offline settings. 

\rev{When deciding how many eigenvectors to compute offline, some applications may be limited by offline time available and by memory constraints. For the $2$D results presented here, offline computation typically took between $1$ and $2$ seconds per eigenvector, for total run-times of less than $5$ minutes. For the $3$D results, offline computation typically took between $1$ to $5$ minutes per eigenvector, and thus the offline run-times could be on the order of several days for large images. In these cases, the \fast RW algorithms presented here may not be useful, dependent on the time between image acquisition and analysis.

Another concern is the memory constraints imposed by using a thousand eigenvectors for $3$D images. Fortunately, the techniques presented do not require all of the eigenvectors to be in memory simultaneously, but rather allow them to be loaded into memory sequentially. This can be used to control the memory usage at a small cost in run-time.}

While the results shown here demonstrate the potential effectiveness of our techniques, the popularity of the RW algorithm has lead to other interesting and potentially powerful extensions that are not discussed here, such as the inclusion of learnt shape priors and automatic seeding for RWIS \cite{baudin2012manifold,baudin2012automatic} and adaptive displacement labels for RWIR \cite{tang2013random}. Further, the RW algorithm was shown by Couprie et al. to be part of a family of graph-based algorithms \cite{couprie2011power}, including other popular algorithms such as Graph Cuts \cite{boykov2001interactive}. Random walks have also been applied to other imaging tasks, such as stereo matching \cite{shen2008stereo} and shape correspondence \cite{oh2012probabilistic}. Our future work will focus on extending our precomputation techniques to the extensions of the RW algorithm, similar graph-based algorithms, and other imaging applications mentioned above.

\bibliographystyle{IEEEtran}
\bibliography{paper_list}

\end{document}